%% file: sigconf.tex
\documentclass[sigconf]{acmart}
\AtBeginDocument{%
  }

\usepackage{multirow}
\usepackage{cleveref}
\usepackage{comment}
\usepackage{threeparttable}
\usepackage{enumitem}
\iffalse
\newcommand{\steve}[1]{{\color{blue}#1}}
\newcommand{\warning}[1]{{\color{red}#1}}
\else
\newcommand{\steve}[1]{#1}
\newcommand{\warning}[1]{}
\fi

\crefname{equation}{Eq.}{Eqs.}
\crefname{figure}{Fig.}{Figs.}

\acmConference[XXX]{XXX}{xxx}{xxx}

\begin{document}

\setlength{\textfloatsep}{4pt plus 2pt minus 4pt}
\setlength{\intextsep}{4pt plus 2pt minus 2pt}

\title{CARROT: Competitive Adaptive Graph Clustering via Semi-Relaxed Optimal Transport}

\author{Ching Fan Chiang}
\affiliation{%
  \institution{National Yang Ming Chiao Tung University}
  \city{Hsinchu}
  \country{Taiwan}}
\email{chiara0105.cs12@nycu.edu.tw}

\author{Jia-Yu Pan}
\affiliation{%
  \institution{Google Research}
  \city{Mountain View}
  \state{CA}
  \country{USA}}
\email{jypan@google.com}

\author{Wen-Chieh Lin}
\affiliation{%
  \institution{National Yang Ming Chiao Tung University}
  \city{Hsinchu}
  \country{Taiwan}}
\email{wclin@cs.nycu.edu.tw}

\renewcommand{\shortauthors}{Fan Chiang et al.}

\begin{abstract}
\steve{Deep graph clustering (DGC) has emerged as a prominent paradigm for unsupervised attributed graph clustering. However, existing methods often struggle with two practical challenges encountered in real-world applications: the number of clusters $K$ is unknown a priori, and cluster sizes are frequently highly imbalanced. Methods relying on a pre-specified $K$ may fail to recover the underlying clustering structure when $K$ is misspecified, while many approaches implicitly favor balanced partitions, hindering the discovery of minority clusters. We propose \textbf{CARROT}, a \textbf{C}ompetitive \textbf{A}daptive g\textbf{R}aph clustering framework via semi-\textbf{R}elaxed \textbf{O}ptimal \textbf{T}ransport, which introduces semi-relaxed optimal transport into deep graph clustering to jointly address both challenges. CARROT first pre-trains node embeddings to establish a meaningful embedding space, and then jointly optimizes embeddings and cluster assignments through an Adaptive Clustering Mechanism. The mechanism combines semi-relaxed optimal transport to adapt cluster capacities to the underlying data, a competition loss to concentrate mass onto well-supported cluster prototypes, thereby starving degenerate ones, and hard pruning to progressively eliminate them. Extensive experiments on five public benchmarks demonstrate that CARROT achieves state-of-the-art NMI and ARI, accurately recovers the number of clusters without supervision, and faithfully preserves size-imbalanced community structures.}
\end{abstract}



\maketitle

\input{Sections/1.Introduction}

\input{Sections/2.Relatedwork}

\input{Sections/3.Preliminary}

\input{Sections/4.Methodology}

\input{Sections/5.Experiments}
\input{Sections/6.Conclusion}




\bibliographystyle{ACM-Reference-Format}
\bibliography{sigconf}

\appendix
\input{Sections/appendix_A}
\input{Sections/appendix_B}

\end{document}

%% file: Sections/1.Introduction.tex
\section{Introduction}
\label{sec:intro}

Graphs are a natural abstraction for representing complex relational systems, spanning social networks, citation graphs, protein interaction networks, and knowledge bases. A fundamental analytical task on such structures is community detection, or graph clustering, which aims to partition nodes into groups whose members are more similar to each other than to the rest of the network, often reflected as denser internal connections than external ones~\cite{def_fortunato2010community, def_fortunato202220years, baseline_blocker2024neuromap}.

When nodes carry rich attribute information, such as user profiles, paper keywords, or biological features, the task becomes attributed graph clustering, where a \textit{good community} must be simultaneously topologically cohesive and semantically coherent. These two objectives, however, are not always aligned. For example, in a co-authorship network where academic disciplines serve as ground-truth labels, two professors sharing the same discipline may have never collaborated, so grouping them together improves semantic coherence but weakens topological cohesion. \steve{The goal of attributed graph clustering is therefore to recover communities that best reconcile these two objectives.}

Deep Graph Clustering (DGC) has emerged as a prominent paradigm for attributed graph clustering, leveraging the representation power of Graph Neural Networks (GNNs)~\cite{gnn_wu2020comprehensive} to jointly optimize node embeddings and cluster assignments in an unsupervised manner~\cite{dgc_liu2026survey, dgc_su2022comprehensive}. 
\steve{Despite their impressive empirical performance, existing DGC methods systematically overlook two practical challenges encountered in real-world graphs.}

First, the number of clusters $K$ is unknown a priori. \steve{For evaluation convenience, most existing works assume that $K$ equals the number of ground-truth classes~\cite{dgc_wang2019daegc, dgc_bo2020sdcn, dgc_yang2023ccgc},} which is inherently impractical in real-world deployment. More fundamentally, methods that require $K$ as a fixed input cannot adapt their partition granularity, making them structurally incapable of discovering the true community structure when $K$ is misspecified.

\steve{Second, real-world communities are often highly imbalanced in size. For example, in the Coauthor PHY dataset~\cite{data_shchur2018pitfalls}, the largest community contains 50.5\% of the nodes, whereas the smallest contains only 7.9\%, a more than sixfold disparity. However, most existing DGC methods are not designed to recover such imbalanced structures~\cite{chall_wang2024dagc}. Without explicit supervision, jointly optimizing embeddings and cluster assignments can lead to cluster collapse, where most or all nodes are assigned to a single cluster~\cite{dgc_tsitsulin2023dmon, chall_ning2026deep}. To avoid this degeneracy, many methods impose strong uniformity-inducing regularization~\cite{dgc_bianchi2020mincutpool, dgc_tsitsulin2023dmon}, which discourages collapse but also suppresses genuine differences in community size. Methods that instead rely on K-means for post-hoc assignment~\cite{dgc_cui2020age, dgc_hou2022graphmae, dgc_liu2022sublime} inherit a similar bias, as K-means tends to produce artificially balanced clusters~\cite{chall_lu2019selfkmeans, chall_He2024ImbalancedDCekm}. Consequently, existing approaches tend to either over-merge minority communities or fragment dominant ones due to an implicit bias toward uniform cluster sizes.
}

\steve{Both challenges call for a cluster assignment mechanism that is both flexible in the number of clusters and adaptive to varying community sizes. 
We argue that Optimal Transport (OT) provides a principled foundation for cluster assignment by formulating it as a global optimization problem.
Rather than assigning nodes independently, OT computes a transport plan that globally minimizes the assignment cost over all node--prototype pairs, allowing assignment decisions to be optimized collectively.

Building on this insight, we propose an Adaptive Clustering Mechanism that introduces semi-relaxed OT into deep graph clustering by integrating it with a competition loss and hard pruning, thereby jointly addressing the issues of unknown $K$ and community size imbalance.
At its core, semi-relaxed OT naturally captures the asymmetry between nodes, which must always be fully assigned, and cluster prototypes, whose capacities should adapt to the underlying community structure rather than be fixed \emph{a priori}. We instantiate the proposed mechanism in \textbf{CARROT}, a \textbf{C}ompetitive \textbf{A}daptive g\textbf{R}aph clustering framework via semi-\textbf{R}elaxed \textbf{O}ptimal \textbf{T}ransport. 
CARROT first learns a meaningful embedding geometry through GNN pre-training, and then jointly optimizes node embeddings and prototype assignments, unifying representation learning and adaptive community discovery in a fully unsupervised, end-to-end manner.}

\steve{
We conduct extensive experiments on five public benchmarks to evaluate CARROT. The results demonstrate that CARROT achieves state-of-the-art clustering performance while accurately recovering the underlying number of communities, producing partitions that are both topologically cohesive and semantically coherent. Qualitative analyses further show that CARROT faithfully preserves size-imbalanced community structures, and sensitivity studies confirm its robustness to the choice of $K_{\max}$, the maximum number of candidate clusters, requiring only a loose upper bound.}

\steve{
The main contributions of this work are summarized as follows:
\begin{itemize}[leftmargin=*, topsep=2pt, itemsep=2pt, parsep=0pt, partopsep=0pt]
    \item 
    A novel semi-relaxed optimal transport formulation for attributed graph clustering, introducing semi-relaxed OT into deep graph clustering.
    \item CARROT, an unsupervised attributed graph clustering framework with an Adaptive Clustering Mechanism that jointly addresses the issues of unknown $K$ and community size imbalance. 
    \item Extensive experiments demonstrating state-of-the-art clustering performance and the robustness of CARROT.
\end{itemize}
}

%% file: Sections/2.Relatedwork.tex
\vspace{-0.5em}
\section{Related Work}
\label{sec:relatedwork}

\subsection{Deep Graph Clustering (DGC)}

Early graph clustering methods, including spectral clustering~\cite{early_shi2000normalizedcuts, early_ng2001spectral, early_von2007tutorial} and stochastic blockmodels~\cite{early_nowicki2001estimation, early_karrer2011stochastic, early_airoldi2008mixed}, operate purely on network structure, leaving node attributes unexploited and scaling poorly to large graphs~\cite{dgc_su2022comprehensive}. 

Deep graph clustering (DGC) addresses these limitations by jointly learning node representations and cluster assignments with Graph Neural Networks (GNNs)~\cite{dgc_liu2026survey, dgc_su2022comprehensive}. Despite their architectural diversity, existing DGC methods require the number of clusters $K$ to be specified in advance~\cite{dgc_wang2019daegc, dgc_bo2020sdcn, dgc_yang2023ccgc, dgc_liu2023hsan, dgc_ren2025mhgc}. 
\steve{To prevent cluster collapse, some methods additionally impose uniform size priors as an inductive bias~\cite{dgc_tsitsulin2023dmon, dgc_bianchi2020mincutpool}, limiting their ability to recover size-imbalanced communities.}  

A growing body of work seeks to relax the requirement for a pre-specified $K$. Among methods that still require an upper bound on the number of clusters, DAG~\cite{baseline_liu2024dag} and DeSE~\cite{baseline_zhang2025dese} set this bound with direct reference to the true number of communities, implicitly encoding prior knowledge of $K$ into the hyperparameter. Neuromap~\cite{baseline_blocker2024neuromap} instead derives the upper bound from the graph structure itself using the Minimum Description Length principle, avoiding prior knowledge of $K$. DGCluster~\cite{baseline_bhowmick2024dgcluster} and GSEC~\cite{baseline_wang2026gsec}, originally proposed for semi-supervised learning, optimize modularity-based objectives without requiring an upper bound on the number of clusters, but neither explicitly considers community size imbalance. 
\steve{ASIL~\cite{dgc_sun2026asil} jointly addresses unknown cluster number and community size imbalance by minimizing structural entropy over a hyperbolic partitioning tree. However, it requires a tighter upper bound on the number of clusters, whereas CARROT requires only a loose upper bound and is robust to this choice. We additionally provide a 
systematic empirical comparison of graph clustering methods that relax the requirement for a pre-specified $K$ in \Cref{subsec:main_results}.}

\vspace{-0.5em}
\subsection{Optimal Transport for Graph Clustering}
Optimal Transport (OT)~\cite{ot_villani2009optimal, ot_peyre2019computational} seeks a minimum-cost transport plan between probability distributions subject to marginal constraints. While the classical formulation requires solving a linear program with cubic complexity in the number of support points, Cuturi~\cite{ot_cuturi2013sinkhorn} introduced entropic regularization, enabling efficient computation via Sinkhorn iterations and making OT tractable for large-scale, end-to-end differentiable learning.

Existing OT-based graph clustering methods adopt standard OT, which enforces hard constraints on both source and target marginals, thereby fixing cluster capacities. \steve{Xu et al.~\cite{ot_xu2019sgwl} formulate graph partitioning as a Gromov--Wasserstein transport problem. Yang et al.~\cite{ot_yang2020otgnn} incorporate entropic OT into unsupervised GNN training for assigning nodes to cluster prototypes. Consequently, both methods inherit this fixed-capacity assumption and are therefore ill-suited for graphs with size-imbalanced communities.}

Unbalanced OT~\cite{ot_chizat2018scaling_uot} relaxes both source and target marginals through KL-divergence penalties, allowing transported mass to deviate from prescribed distributions. For graph clustering, however, relaxing the source marginal is undesirable since some nodes may contribute no mass to any cluster prototype, leaving them effectively unassigned. Semi-relaxed OT instead relaxes only the target marginal while preserving the source constraint, ensuring every node is fully assigned while allowing prototype capacities to adapt. 

Vincent-Cuaz et al.~\cite{ot_VincentCuaz2022SemirelaxedGD} exploit this asymmetric formulation in the Gromov--Wasserstein setting for graph partitioning and completion. However, Gromov-Wasserstein transport compares intra-graph pairwise distance structures across two spaces rather than computing transport costs between points in a shared embedding space.
To the best of our knowledge, no prior work has applied semi-relaxed OT to node-to-prototype assignment for deep graph clustering, a gap that CARROT addresses.

%% file: Sections/3.Preliminary.tex
\section{Preliminary}
\label{sec:preliminary}
\steve{We first formulate the problem and then introduce the optimal transport tools underlying CARROT’s assignment procedure.}

\subsection{Problem Formulation}
Let $\mathcal{G} = (\mathcal{V}, \mathcal{E}, \mathbf{X})$ be an attributed graph, where $\mathcal{V}=\{v_1,\dots,v_N\}$ is a set of $N$ nodes, $\mathcal{E}$ is the edge set, and $\mathbf{X}\in\mathbb{R}^{N\times D}$ is the node feature matrix with row $\mathbf{x}_i\in\mathbb{R}^D$ representing the attribute vector of node $v_i$. \steve{The goal of attributed graph clustering is to partition $\mathcal{V}$ into $K$ disjoint clusters based on both graph structure and node features, where $K$ is unknown a priori. A desirable partition exhibits both topological cohesion, with denser intra-cluster than inter-cluster connections, and semantic coherence, grouping nodes with similar attributes together.}

We frame cluster assignment as a mass-transport problem, treating the set of nodes as a discrete source distribution to be transported onto $K$ learnable cluster prototypes acting as target locations in the embedding space.

\steve{
\subsection{Optimal Transport (OT)}
Given two discrete distributions $\mathbf{a} \in \mathbb{R}_+^N$ and $\mathbf{b} \in \mathbb{R}_+^K$ over $N$ source points $\{\mathbf{s}_i\}_{i=1}^{N}$ and $K$ target points $\{\mathbf{t}_j\}_{j=1}^{K}$, respectively, Optimal Transport (OT) seeks a minimum-cost plan for transporting mass from $\mathbf{a}$ to $\mathbf{b}$: 
\begin{equation}
\min_{\mathbf{P} \in U(\mathbf{a}, \mathbf{b})} \langle \mathbf{P}, \mathbf{C} \rangle,
\label{eq:OTobjective}
\end{equation}
where $\mathbf{P} \in \mathbb{R}_+^{N \times K}$ is the transport plan, $\mathbf{C} \in \mathbb{R}_+^{N \times K}$ is the transportation cost matrix, and $\langle \cdot, \cdot \rangle$ denotes the Frobenius inner product. 

Each entry $P_{ij}$ specifies the amount of mass transported from source point $\mathbf{s}_i$ to target point $\mathbf{t}_j$, while $C_{ij}$ denotes the corresponding unit transportation cost. $U$ is the set of feasible transport plans forming a \textit{transportation polytope}:
\begin{equation}
U(\mathbf{a}, \mathbf{b}) = \{ \mathbf{P} \in \mathbb{R}_+^{N \times K} \mid \mathbf{P}\mathbf{1}_K = \mathbf{a},\; \mathbf{P}^\top\mathbf{1}_N = \mathbf{b} \},
\end{equation}
where $\mathbf{1}_K \in \mathbb{R}^K$ and $\mathbf{1}_N \in \mathbb{R}^N$ denote all-ones vectors. The constraint $\mathbf{P}\mathbf{1}_K=\mathbf{a}$ ensures mass conservation by requiring every source point to distribute all of its mass. The constraint $\mathbf{P}^{\top}\mathbf{1}_N=\mathbf{b}$ fixes the amount of mass assigned to each target point according to the prescribed target marginal. 

Adding an entropic regularization term $\epsilon H(\mathbf{P})$ to \Cref{eq:OTobjective} yields a strongly convex objective with a unique solution $\mathbf{P}^*$ that varies differentiably with $\mathbf{C}$ and can be computed efficiently using the Sinkhorn-Knopp algorithm~\cite{ot_cuturi2013sinkhorn}. $\epsilon > 0$ controls the regularization strength; a larger $\epsilon$ produces softer assignments, while a smaller $\epsilon$ recovers a sharper plan closer to the unregularized solution. The entropy term is defined as 
\begin{equation} 
H(\mathbf{P}) = -\sum_{ij} P_{ij} \log P_{ij}. \end{equation}
}

\steve{
\subsection{Semi-Relaxed Optimal Transport}
\label{subsec:preliminary_srOT}
Semi-Relaxed OT relaxes the target marginal constraint of OT by treating the target marginal $\mathbf{b}$ as a soft prior rather than a hard constraint. This flexibility is particularly beneficial for graph clustering, where cluster sizes are generally unknown and often highly imbalanced. Instead of enforcing $\mathbf{P}^{\top}\mathbf{1}*N=\mathbf{b}$ exactly, semi-relaxed OT penalizes deviations from $\mathbf{b}$ using a KL divergence term:
\begin{equation}
\min_{\mathbf{P} \geq 0} \langle \mathbf{P}, \mathbf{C} \rangle - \epsilon H(\mathbf{P}) + \rho\, \mathrm{KL}(\mathbf{P}^\top \mathbf{1}_N \,\|\, \mathbf{b}) \quad \text{s.t.} \quad \mathbf{P}\mathbf{1}_K = \mathbf{a},
\end{equation}
where $\epsilon>0$ controls the entropic regularization, $\rho>0$ controls the strength of the target marginal constraint, and $\mathbf{b}\in\mathbb{R}_+^K$ denotes the target prior.

The optimal transport plan can be computed efficiently using a generalized Sinkhorn--Knopp procedure~\cite{ot_cuturi2013sinkhorn}. Specifically, the optimal plan has the form:
\begin{equation}
\mathbf{P}^* = \mathrm{diag}(\mathbf{u})\, \mathbf{K}\, \mathrm{diag}(\mathbf{v}), \quad \text{where} \quad K_{ij} = \exp(-C_{ij}/\epsilon),
\end{equation}
where $\mathbf{u}\in\mathbb{R}_+^N$ and $\mathbf{v}\in\mathbb{R}_+^K$ are scaling vectors, and $\mathrm{diag}(\mathbf{u})$ denotes the diagonal matrix with diagonal entries given by $\mathbf{u}$. Let $\tau=\rho/(\rho+\epsilon)$ denote the relaxation factor. The scaling vectors are updated iteratively as
\begin{equation}
\mathbf{u}
\leftarrow
\frac{\mathbf{a}}{\mathbf{K}\mathbf{v}},
\qquad
\mathbf{v}
\leftarrow
\left(
\frac{\mathbf{b}}{\mathbf{K}^{\top}\mathbf{u}}
\right)^{\tau},
\label{eq:Sinkhorn_update}
\end{equation}
where all divisions and exponentiation are applied element-wise. As $\rho\rightarrow\infty$, $\tau\rightarrow1$, recovering the standard Sinkhorn--Knopp updates for OT. For finite $\rho$, $\tau<1$ relaxes the target marginal constraint, allowing the transport plan to naturally adapt to imbalanced cluster sizes. The specific choices of the source marginal $\mathbf{a}$ and target prior $\mathbf{b}$ are described in \Cref{subsubsec:method_srOT_design}.
}

%% file: Sections/4.Methodology.tex
\section{Methodology}
\label{sec:method}

\begin{figure}[h] 
    \centering
    \includegraphics[width=\columnwidth]{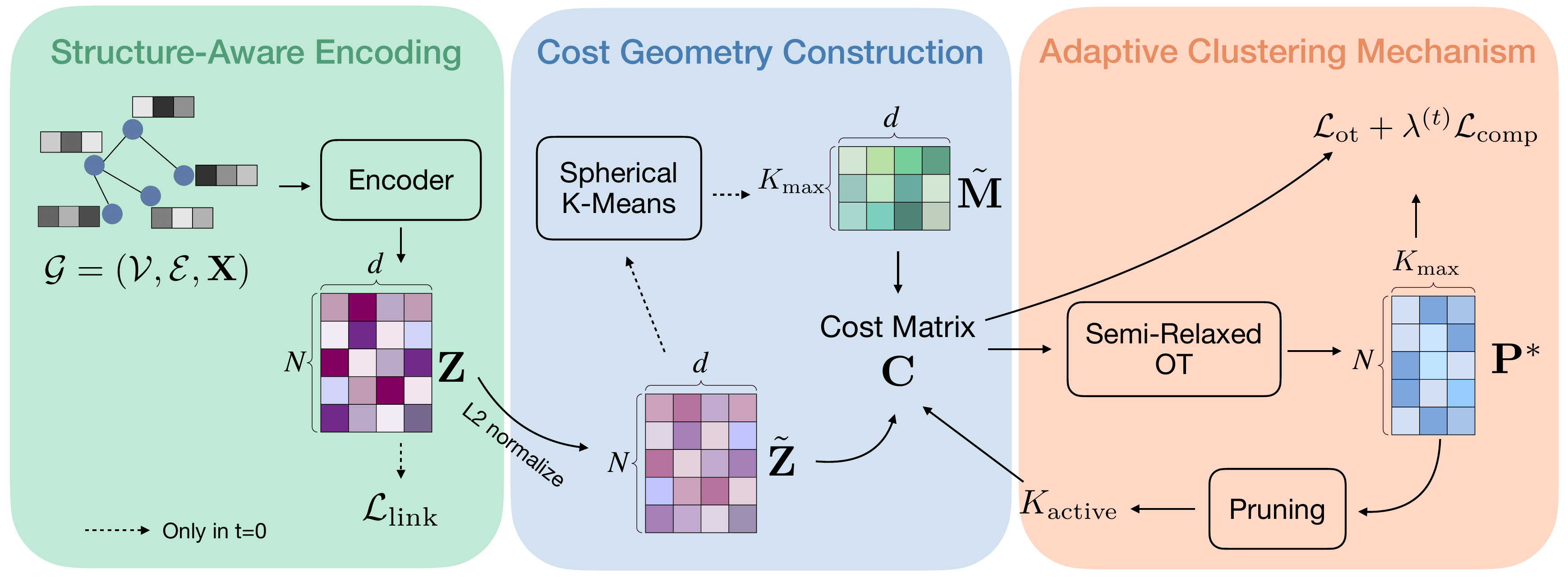}
    \caption{The overall framework of CARROT.}
    \label{fig:framework}
\end{figure}

\steve{CARROT consists of three components. As illustrated in \Cref{fig:framework}, \textbf{Structure-Aware Encoding} (\Cref{subsec:method_encoding}) first pre-trains node embeddings to establish a geometrically meaningful embedding space grounded in graph topology. \textbf{Cost Geometry Construction} (\Cref{subsec:method_costgeom}) then embeds nodes and prototypes into a shared geometry from which the transport cost is defined. Built upon these components, the proposed \textbf{Adaptive Clustering Mechanism} (\Cref{subsec:method_acm}) jointly addresses unknown $K$ and community size imbalance without supervision. Finally, the resulting objective is optimized end-to-end to jointly optimize the GNN encoder and the prototype matrix $\mathbf{M}$ (\Cref{subsec:method_optimization}).}

\subsection{Structure-Aware Encoding}
\label{subsec:method_encoding}

\steve{To establish a stable geometric foundation for clustering, CARROT first pre-trains a GNN encoder using self-supervised contrastive learning. This stage produces embeddings that preserve the graph topology before prototype optimization begins.}

\textbf{GNN Encoder.}
\steve{We employ a two-layer Graph Convolutional Network (GCN)~\cite{gnn_kipf2017semisupervised} with a ReLU activation between layers to map each node to a $d$-dimensional embedding:}
\begin{equation}
    \mathbf{Z} = \text{Encoder}(\mathbf{X}, \mathcal{E}).
\end{equation}

\textbf{Link Prediction Loss.}
\steve{The encoder is pre-trained using a self-supervised link prediction objective~\cite{gnn_kipf2016vgae}. The objective encourages connected nodes to have similar embeddings while separating disconnected nodes, thereby preserving the graph topology in the learned embedding space:}
\begin{equation}
    \mathcal{L}_{\text{link}} = -\mathbb{E}_{(u,v)}\left[ y_{uv} \log \sigma(\langle \mathbf{z}_u, \mathbf{z}_v \rangle) + (1 - y_{uv})\log(1 - \sigma(\langle \mathbf{z}_u, \mathbf{z}_v \rangle)) \right],
\end{equation}
where $y_{uv}=1$ if $(u,v)\in\mathcal{E}$ and $0$ otherwise, and $\sigma(\cdot)$ denotes the sigmoid function. \steve{At each training epoch, negative edges are sampled uniformly from non-adjacent node pairs, with the number of negative samples matched to that of the positive edges.}

\subsection{Cost Geometry Construction}
\label{subsec:method_costgeom}

We next construct a shared geometry between node embeddings and prototypes, from which the transport cost for cluster assignment is defined.

\textbf{Embedding Normalization and Cost Matrix.}
At the start of each epoch of joint optimization, we apply L2-normalization to each node embedding and each prototype vector, projecting both onto the unit hypersphere, yielding $\tilde{\mathbf{Z}}$ and $\tilde{\mathbf{M}}$. \steve{This normalization provides three benefits.} 
First, GCN aggregation assigns larger embedding norms to high-degree nodes, introducing a degree-induced bias into raw representations; row-wise normalization removes this bias, ensuring that similarity is governed by the directional structure of embeddings rather than their magnitudes. \steve{Second, on the unit hypersphere, squared Euclidean distance decreases linearly with cosine similarity. We therefore define the transport cost matrix $\mathbf{C}\in\mathbb{R}_+^{N\times K_{\max}}$ as}
\begin{equation}
    C_{ij} = \|\tilde{\mathbf{z}}_i - \tilde{\mathbf{m}}_j\|^2,
\end{equation}
so that minimizing transport cost is equivalent to maximizing cosine similarity between nodes and their assigned prototypes. Third, all pairwise distances are bounded within $[0, 4]$, ensuring that the Gibbs kernel $\mathbf{K} = \exp(-\mathbf{C}/\epsilon)$ remains numerically stable across datasets and that $\epsilon$ does not require dataset-specific tuning.

\textbf{Prototype Initialization.}
After pre-training, we initialize the prototype matrix $\mathbf{M}\in\mathbb{R}^{K_{\max}\times d}$ by applying Spherical K-means to the normalized node embeddings $\tilde{\mathbf{Z}}$, setting each row of $\mathbf{M}$ to a cluster centroid:
\begin{equation}    \mathbf{M}=\text{SphericalKMeans\_Centroids}(\tilde{\mathbf{Z}},\,K_{\max}).
\end{equation}
\steve{
Since Spherical K-means produces normalized centroids, the initialized prototypes ($\mathbf{M}^{(0)} = \tilde{\mathbf{M}}^{(0)}$) are already consistent with the normalized embedding space used throughout optimization. Moreover, because Spherical K-means minimizes cosine distance under the same geometry, it provides a geometrically consistent starting point for the subsequent joint optimization.}

\subsection{Adaptive Clustering Mechanism}
\label{subsec:method_acm}

The Adaptive Clustering Mechanism jointly optimizes the GNN encoder and prototype matrix $\mathbf{M}$. The mechanism consists of three tightly coupled components that form a sequential optimization pipeline. Semi-relaxed OT first relaxes the target marginal, permitting mass to distribute across prototypes according to the cost geometry rather than a predefined uniform prior (\Cref{subsubsec:method_srOT_design}).
The competition loss then drives rivalry among prototypes, concentrating mass onto well-supported prototypes while starving degenerate ones (\Cref{subsubsec:method_loss_weighting}).
Hard pruning subsequently removes starved prototypes, reducing $K_{\text{active}}$ and updating the masking to block future mass flow (\Cref{subsubsec:method_pruning_masking}).
\steve{Training begins with $K_{\max}$ active prototypes, where $K_{\max}$ serves as a loose upper bound on the cluster count. Throughout optimization, CARROT progressively eliminates degenerate prototypes, reducing $K_{\text{active}}$ until convergence.}

\subsubsection{Semi-Relaxed OT Assignment}
\label{subsubsec:method_srOT_design}
\steve{
\mbox{}\\
\textbf{Marginal Specification.} We set the source marginal to $\mathbf{a}=\mathbf{1}_N$, assigning one unit of transport mass to each node. Each node can distribute this mass across multiple active prototypes, with $\sum_j P^*_{ij}=1$ ensuring that the resulting assignments form a probability distribution. Accordingly, $P^*_{ij}$ denotes the soft-assignment probability of node $i$ to prototype $j$. 

For the target prior, we assign $b_j=N/K_{\text{active}}$ for active prototypes and a negligible nominal value $b_j=\delta$ for inactive ones to keep the $\mathbf{v}$-updates (\Cref{eq:Sinkhorn_update}) numerically well-defined, where $K_{\text{active}}$ is the current number of non-pruned prototypes. The target marginal $\mathbf{b}$ is recomputed after each pruning step to reflect the updated set of active prototypes. Inactive prototypes are additionally masked in the cost matrix to prevent further mass assignment.

This design provides a neutral prior that defaults to uniform assignments in the absence of structural evidence, yet allows the transport plan to deviate freely whenever the learned cost geometry supports imbalanced communities, unlike the hard uniformity constraints adopted by prior methods.}

\textbf{Sinkhorn Solver.}
Given the cost matrix $\mathbf{C}$, marginals $(\mathbf{a},\mathbf{b})$, entropic \steve{regularization strength} $\epsilon$, and unbalanced penalty \steve{strength} $\rho$, we compute the optimal transport plan $\mathbf{P}^*$ using the generalized Sinkhorn iterations described in \Cref{subsec:preliminary_srOT}. The resulting transport plan serves as the soft cluster assignment matrix for the subsequent \steve{optimization}.

\subsubsection{Loss Functions and Dynamic Loss Weighting}
\label{subsubsec:method_loss_weighting}
\mbox{}\\
\textbf{OT Alignment Loss.} This objective minimizes the total transport cost between node embeddings and their assigned prototypes:
\begin{equation}
    \mathcal{L}_{\text{ot}}=\frac{1}{N}\left\langle\mathbf{P}^*, \mathbf{C}\right\rangle=\frac{1}{N} \sum_{i, j} P_{i j}^* C_{i j}.
\end{equation}
\steve{Optimizing $\mathcal{L}_{\text{ot}}$ jointly pulls node embeddings toward their assigned prototypes while updating prototypes toward the weighted centroid of their assigned nodes. The $1/N$ normalization makes the loss scale invariant to the dataset size.}

\textbf{Competition Loss.}
To drive the adaptive clustering mechanism, we propose a competition loss that encourages mass to concentrate on a small number of representative prototypes. Let $\pi_j = \frac{1}{N}\sum_{i=1}^N P^*_{ij}$ denote the mass fraction absorbed by \steve{prototype $j$}. 
The competition loss is defined as the negative squared L2-norm of $\boldsymbol{\pi}$:
\begin{equation}
    \mathcal{L}_{\text{comp}} = -\sum_{j=1}^{K_{\max}} \pi_j^2=-\|\boldsymbol{\pi}\|_2^2.
\end{equation}
\steve{By the power-mean inequality, $\|\boldsymbol{\pi}\|_2^2$ is maximized when mass concentrates on fewer prototypes and minimized under a uniform mass distribution.} Minimizing $\mathcal{L}_{\text{comp}}$ thus creates a competitive pressure among prototypes: \steve{well-supported prototypes attract more mass while starving weak ones, making them candidates for pruning.}

\textbf{Dynamic Loss Weighting.}
The magnitudes of $\mathcal{L}_{\text{ot}}$ and $\mathcal{L}_{\text{comp}}$ evolve differently throughout training, so a fixed weighting cannot maintain a consistent balance between them. We therefore propose an adaptive loss-weighting scheme that continuously rescales the competition loss according to
\begin{equation}
    \lambda^{(t)} = \frac{r \cdot \mathcal{L}_{\text{ot}}^{(t-1)}}{\mathcal{L}_{\text{comp}}^{(t-1)}}.
\end{equation}
This design maintains competition pressure proportional to the current alignment signal throughout training. Early in optimization, a larger $\mathcal{L}_{\text{ot}}$ increases the influence of $\mathcal{L}_{\text{comp}}$ to promote prototype differentiation. As optimization converges, the weight naturally decreases, preventing excessive mass concentration after the partition has stabilized.

\subsubsection{Inactive Cluster Pruning and Masking}
\label{subsubsec:method_pruning_masking}

\steve{
The competition loss drives mass to concentrate on representative prototypes, but concentration alone does not reduce $K$. Hard pruning bridges this gap, converting this soft competitive pressure into discrete structural updates by progressively reducing $K_{\text{active}}$.}

\textbf{Pruning Criterion.}
Every $T_{\text{prune}}$ epochs, each active prototype is evaluated by its hard assignment count under the current $\mathbf{P}^*$. Prototypes with fewer than $\alpha N/K_{\text{active}}$ assigned nodes, where $\alpha\in(0,1)$ is a sensitivity hyperparameter, are marked as inactive, reducing $K_{\text{active}}$ accordingly.

\textbf{Inactive Prototype Masking.}
Pruned prototypes are retained in $\mathbf{M}$ to preserve dimensional consistency during backpropagation, but are excluded from subsequent assignments by masking their corresponding cost-matrix columns with a prohibitively large constant. This drives the associated Sinkhorn kernel entries \steve{$K_{ij}$} to zero, \steve{preventing further mass assignment.} Pruned prototypes therefore neither attract mass nor contribute to any loss computation in subsequent epochs.
The hyperparameters $T_{\text{prune}}$, $\alpha$, and the sentinel value used for masking are provided in Appendix~\ref{app:hyperparameters}.

\subsection{Optimization}
\label{subsec:method_optimization}
The overall training objective combines the OT alignment and competition losses using the dynamic weight:
\begin{equation}
    \mathcal{L} = \mathcal{L}_{\text{ot}} + \lambda^{(t)} \mathcal{L}_{\text{comp}}.
    \label{eq:TotalLoss}
\end{equation}
The Sinkhorn iterations produce a transport plan $\mathbf{P}^*$ that is differentiable with respect to the cost matrix $\mathbf{C}$ (\Cref{subsec:preliminary_srOT}), and hence with respect to both the node embeddings and the prototypes. Gradients therefore propagate end-to-end through $\mathbf{P}^*$ and $\mathbf{C}$ to jointly optimize the GNN encoder and prototype matrix $\mathbf{M}$.

We further adopt a dual learning-rate scheme, using a substantially smaller learning rate for the GNN encoder than for $\mathbf{M}$. This reflects their complementary roles: the pre-trained encoder provides a stable geometric foundation, whereas the prototypes must adapt rapidly to the evolving embedding space throughout clustering. The learning rates for both components are provided in Appendix~\ref{app:hyperparameters}. Time complexity is analyzed in Appendix~\ref{app:complexity}.

%% file: Sections/5.Experiments.tex
\section{Experiments}
\label{sec:experiments}

We evaluate CARROT through quantitative and qualitative experiments. \Cref{subsec:main_results} compares CARROT with competitive baselines on clustering performance. \Cref{subsec:qualitative} visualizes cluster assignments on size-imbalanced datasets to demonstrate CARROT's ability to recover imbalanced community structures. \Cref{subsec:ablation} presents ablation studies of CARROT's key components. Finally, \Cref{subsec:sensitivity} evaluates robustness to $K_{\max}$ and sensitivity to the Sinkhorn hyperparameters $\epsilon$ and $\rho$.

\subsection{Experimental Setup}
\label{subsec:exp_setup}
\textbf{Datasets.} We evaluate CARROT on five widely used attributed graph clustering benchmarks covering citation, co-purchase, and co-authorship networks. As summarized in Table~\ref{tab:datasets}, the datasets vary substantially in graph scale and exhibit different degrees of community size imbalance. Since CARROT is designed to recover clusters with unknown and potentially imbalanced sizes, we include both relatively balanced (PubMed and CiteSeer) and highly imbalanced (Amazon PC and Coauthor PHY) datasets. We quantify cluster-size imbalance using \emph{Size CV}, the coefficient of variation of cluster sizes, defined as the ratio of their standard deviation to their mean, where larger values indicate greater imbalance.
Additional dataset details are provided in Appendix~\ref{app:dataset_details}.
\input{tables/table-datasets}

\noindent\textbf{Evaluation Metrics.} We evaluate clustering quality using both external and internal metrics. The external metrics, Normalized Mutual Information (NMI)~\cite{metric_strehl2002cluster} and Adjusted Rand Index (ARI)~\cite{metric_Hubert1985ComparingP}, measure agreement with ground-truth labels. NMI captures global distributional agreement, whereas ARI measures pairwise agreement between predicted and ground-truth assignments. The internal metrics, Modularity ($Q$)~\cite{early_newman2006modularity} and the Calinski--Harabasz Index (CHI)~\cite{metric_calinski1974dendrite}, assess topological cohesion and semantic compactness, respectively, without requiring ground-truth labels. We also report their product, $Q \times \mathrm{CHI}$, as a combined internal quality measure. We additionally report Found-$K$, the number of clusters produced by each method, to evaluate its ability to infer the number of clusters. All results are averaged over 10 random seeds (0--9) and reported as mean $\pm$ standard deviation.

\steve{
\noindent\textbf{Baselines.}
We compare CARROT against five deep graph clustering methods with adaptive-$K$ capability: DAG~\cite{baseline_liu2024dag}, DeSE~\cite{baseline_zhang2025dese}, Neuromap~\cite{baseline_blocker2024neuromap}, DGCluster~\cite{baseline_bhowmick2024dgcluster}, and GSEC~\cite{baseline_wang2026gsec}. DAG, DeSE, and Neuromap require an upper bound $K_{\max}$ and are evaluated under a unified setting of $K_{\max}=50$ to ensure a fair comparison. 
These methods were selected because we were able to reproduce their results using publicly available code.
For DAG, since its source code is not publicly available, we report the results from the original paper using its original setting ($K_{\max}=2K_{\text{true}}$), denoted by~$\dagger$. To ensure a unified unsupervised evaluation protocol across all methods, we use $Q \times \mathrm{CHI}$ as the model selection criterion whenever checkpoint selection is required~\cite{baseline_liu2024dag}. 
DGCluster and GSEC do not require $K_{\max}$, denoted by~$\ddagger$, and are evaluated using the unsupervised settings reported in their original papers without incorporating label priors.}

\steve{
\noindent\textbf{Experimental settings.}
CARROT uses a GNN encoder that produces node embeddings of dimension $d = 32$ for all datasets. The key algorithmic hyperparameters are shared across all datasets, with the entropic regularization strength set to $\epsilon=0.1$ and the unbalanced penalty set to $\rho=0.05$. Following DAG~\cite{baseline_liu2024dag}, we use $Q \times \mathrm{CHI}$ as the unsupervised checkpoint selection criterion and retain the checkpoint with the highest $Q \times \mathrm{CHI}$ score for evaluation. Complete training configurations and details of the computational environment are provided in Appendices \ref{app:hyperparameters} and \ref{app:environment}, respectively.
}

\input{tables/table-main-results}

\subsection{Main Results}
\label{subsec:main_results}
Table~\ref{tab:main_resluts} reports the clustering results of CARROT against five deep graph clustering baselines across five benchmarks. For all methods, both the graph structure and node attributes provided by each dataset are used as input, ensuring a consistent information setting across comparisons. 

CARROT outperforms all baselines in terms of both NMI and ARI across all five datasets, \steve{demonstrating its ability to jointly address unknown K estimation and community size imbalance.}
We further observe a consistent pattern among \steve{all} baselines: NMI substantially exceeds ARI, with an average gap of $0.13$ across methods and datasets, and reaching as high as $0.25$ on Coauthor PHY (DGCluster), indicating that their partitions capture the global label distribution but fail to achieve precise pairwise agreement across all communities. CARROT is the only method that largely closes this gap: its NMI--ARI difference remains below $0.06$ on all datasets, and ARI actually exceeds NMI on CiteSeer and Coauthor PHY. This balance is attributed to CARROT's semi-relaxed OT assignment, which accommodates heterogeneous cluster sizes and avoids the systematic bias toward majority communities that penalizes ARI more severely than NMI; we isolate and verify this effect directly in the ablation study (\Cref{subsec:ablation}).

\steve{On Found-$K$, CARROT consistently recovers cluster counts close to the ground truth across all five datasets without supervision.}
Methods such as Neuromap and DGCluster consistently over-segment, producing two to four times \steve{more clusters than}  
the ground truth on citation networks, while GSEC tends to under-segment on Amazon PC (Found-$K$$=$$4.9$, $K_{\text{true}}$$=$$10$). CARROT avoids both failure modes across all datasets. \steve{This is enabled by the Adaptive Clustering Mechanism, which jointly drives and validates prototype elimination throughout training.}

\steve{On the internal metrics, CARROT achieves the highest $Q{\times}\mathrm{CHI}$ score on three datasets (Cora, CiteSeer, and PubMed) and ranks second on Coauthor PHY. The only exception is Amazon PC, where GSEC and DGCluster achieve higher scores. However, both methods substantially under-segment the graph relative to the ground truth ($K_{\text{true}} = 10$), and the CHI is known to favor coarser partitions when the number of clusters differs substantially across solutions~\cite{results_chikumbo2019optimal,results_arbelaitz2013extensive}. Consequently, $Q{\times}\mathrm{CHI}$ should be interpreted with caution when comparing methods that recover markedly different numbers of clusters. A more detailed discussion of this limitation of CHI is provided in Appendix~\ref{app:CHI_bias}.}

\begin{figure*}[t] 
    \centering
    \includegraphics[width=\textwidth]{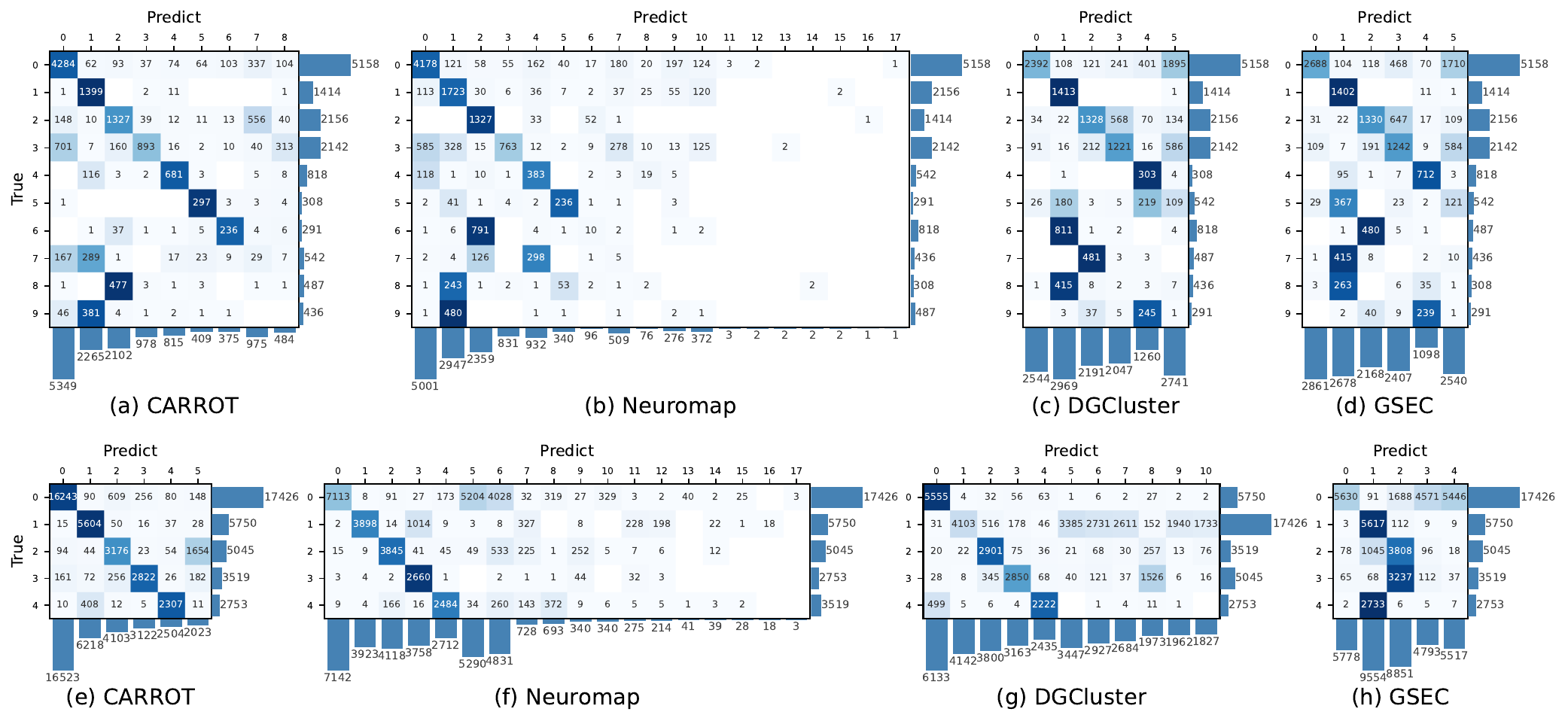}
    \caption{Cluster assignment matrices with marginal histograms on the Amazon PC (a)--(d) and Coauthor PHY (e)--(h) datasets, comparing CARROT, Neuromap, DGCluster, and GSEC. Each cell $(i,j)$ reports the number of nodes from true cluster $i$ assigned to predicted cluster $j$; cell shading indicates the proportion of these nodes within true cluster $i$. \steve{Bottom} 
    marginal histograms show the predicted cluster size distribution; right marginals show the true cluster sizes.}
    \label{fig:two_datasets_heatmap}
\end{figure*}

\subsection{Qualitative Analysis}
\label{subsec:qualitative}

\steve{To gain deeper insight into clustering behavior under class-imbalanced conditions, we visualize the cluster assignment matrices of CARROT, Neuromap, DGCluster, and GSEC on two datasets with pronounced size imbalance: Amazon PC and Coauthor PHY. Since the number of predicted clusters may differ from the ground truth, cluster indices have no cross-method correspondence. Predicted clusters are therefore reordered using Hungarian matching to maximize diagonal occupancy. Each cell reports the number of nodes assigned from a true cluster to a predicted cluster, while cell shading indicates the corresponding within-cluster proportion. Consequently, larger and darker diagonal cells indicate better recovery of the true partition.
}

\steve{On Amazon PC (\Cref{fig:two_datasets_heatmap}(a)--(d); true partition: 10 clusters; largest and smallest cluster: 5,158 and 291 nodes), CARROT recovers 9 clusters and exhibits the greatest concentration of large, dark diagonal cells among all four methods, indicating the highest overall per-cluster recall while preserving meaningful size heterogeneity in its predicted partition. Neuromap predicts substantially more clusters than the true partition size, yet achieves the second-greatest diagonal concentration, suggesting that its dominant predicted clusters still align reasonably well with the true communities. In contrast, DGCluster and GSEC both tend toward uniform cluster sizes, leading to systematic over-merging of small communities. In DGCluster, predicted cluster 1 absorbs three true clusters (1, 6, and 8), while in GSEC, predicted cluster 1 merges four true clusters (1, 5, 7, and 8); in both cases, prominent large and dark off-diagonal cells reflect these incorrect assignments.}

\steve{On Coauthor PHY (\Cref{fig:two_datasets_heatmap}(e)--(h); true partition: 5 clusters; largest and smallest cluster: 17,426 and 2,753 nodes), the true partition is dominated by a single community comprising 50.5\% of all nodes. CARROT recovers this super-majority cluster with 93.2\% recall (16,243 of 17,426 nodes), achieving the highest per-cluster recall among all four methods. Neuromap, DGCluster, and GSEC each fragment this dominant community across multiple predicted clusters, and none exhibits a diagonal pattern comparable to CARROT's. Although GSEC recovers the correct number of clusters, its predicted cluster-size distribution still deviates substantially from the true partition due to simultaneous over-fragmentation of the dominant community and over-merging of smaller ones, producing several prominent off-diagonal cells.}

\steve{Across both datasets, CARROT consistently recovers the size-imbalanced community structure most faithfully, leveraging its semi-relaxed OT formulation to accommodate unequal cluster masses and produce cluster assignments whose diagonal concentration and size distribution closely match the true partition. In contrast, DGCluster and GSEC exhibit systematic over-merging and over-fragmentation, reflecting an implicit preference for more uniform cluster sizes that is ill-suited to these imbalanced datasets. Neuromap avoids this bias but still produces a weaker diagonal pattern than CARROT.}

\vspace{-1em} 
\subsection{Ablation Study}
\label{subsec:ablation}
To evaluate the contributions of the semi-relaxed OT formulation and the competition loss $\mathcal{L}_{\text{comp}}$, we perform two ablations (\Cref{tab:ablation_mean}):  

\textbf{w/o semi-relaxed} -- Replacing semi-relaxed OT with standard OT enforces the target marginal as a hard constraint, fixing the transport plan to the uniform prior regardless of the underlying community structure.
Consequently, Found-$K$ remains close to $K_{\max}$ on all datasets, Size~CV drops substantially, and both NMI and ARI deteriorate markedly. These results indicate that removing the soft relaxation on the target marginal, rather than the uniformity of the prior itself, is what impairs recovery of size-imbalanced community structure.

\textbf{w/o $\mathcal{L}_{\text{comp}}$} -- Removing $\mathcal{L}_{\text{comp}}$ eliminates the competitive pressure among prototypes. As a result, Found-$K$ remains fixed at $K_{\max}=50$ across all datasets, accompanied by substantial degradation in NMI and ARI. Moreover, we empirically observe that the resulting cluster-size distribution closely resembles the Spherical K-means initialization, indicating that semi-relaxed OT alone only refines the assignment at a small scale, insufficient to substantially reshape the cluster-size distribution. This confirms that $\mathcal{L}_{\text{comp}}$ is essential not only for concentrating mass onto representative prototypes, but for providing the driving force needed to move the assignment meaningfully away from its initialization.

The full CARROT model achieves the best NMI and ARI across all five benchmarks, with particularly large gains in ARI, which is more sensitive to whether both the number and boundaries of communities are correctly recovered. These results highlight the importance of both designs: the semi-relaxed formulation grants the freedom to express genuine size imbalance, and $\mathcal{L}_{\text{comp}}$ fosters competition among cluster prototypes to resolve unknown $K$ by eliminating degenerate ones.

\input{tables/table-ablation}

\subsection{Sensitivity Analysis}
\label{subsec:sensitivity}
\begin{figure}[h]
    \centering
    \includegraphics[width=\columnwidth]{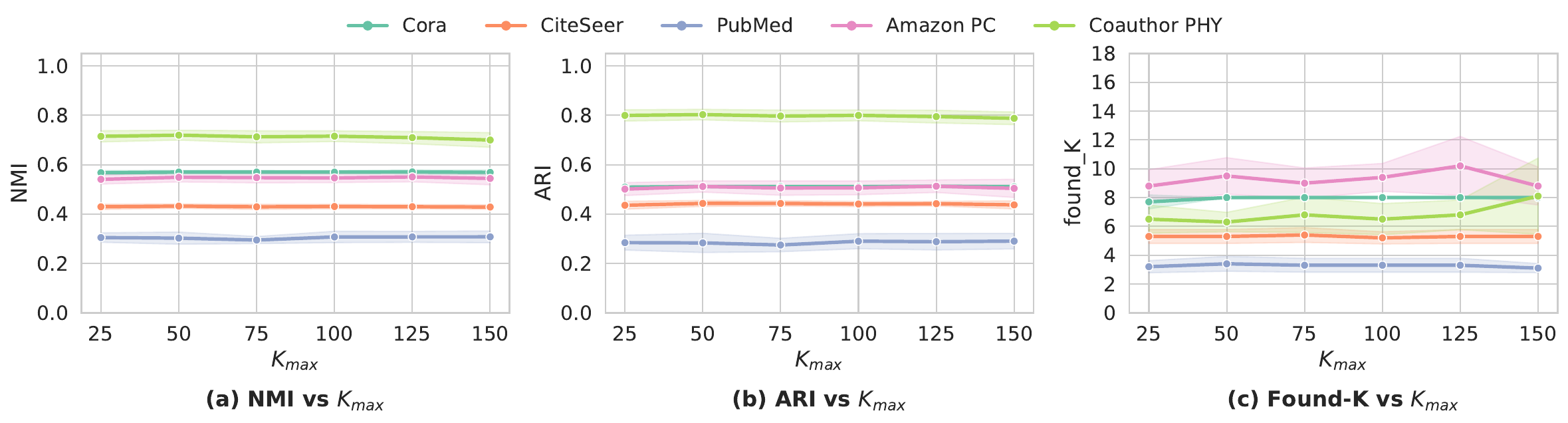}
    \caption{Sensitivity analysis across varying $K_{\max}$: (a) NMI, (b) ARI, (c) Found-K. Lines and shaded bands denote mean and standard deviation across seeds.}
\label{fig:stability_k}
\end{figure}

We evaluate CARROT with $K_{\max} \in \{25, 50, 75, 100, 125, 150\}$ while keeping all other hyperparameters fixed. \Cref{fig:stability_k} reports NMI, ARI, and Found-$K$ averaged over 10 random seeds. Results for Modularity and CHI exhibit the same trends and are provided in Appendix \ref{app:sensitivity_kmax}.

As shown in \Cref{fig:stability_k}, CARROT is robust to the choice of $K_{\max}$. Across all datasets, NMI and ARI fluctuate only within the level of run-to-run noise. Likewise, Found-$K$ remains close to $K_{\text{true}}$, differing by less than one cluster on most datasets and by fewer than two clusters even on Coauthor PHY, which exhibits the largest variation.

These results demonstrate that $K_{\max}$ functions only as a loose upper bound rather than a parameter requiring careful tuning. As long as it is larger than the true number of clusters, CARROT consistently recovers both the underlying community structure and an accurate estimate of the cluster count, substantially reducing the burden of hyperparameter selection in fully unsupervised settings. 

We additionally examine sensitivity to the Sinkhorn hyperparameters $\epsilon$ and $\rho$, finding that CARROT is markedly robust to $\rho$ while $\epsilon$ requires more careful tuning; full results and discussion are provided in Appendix~\ref{app:sensitivity_eps_rho}.

%% file: tables/table-datasets.tex
\begin{table}[h]
\centering
\caption{Dataset statistics.}
\label{tab:datasets}
\small
\begin{tabular}{lrrrrr}
\toprule
Dataset & \#Node & \#Edge & \#Feature & \#Cluster & Size CV \\
\midrule
Cora          &  2,708 &   5,278 & 1,433 &  7 & 0.5069 \\
CiteSeer      &  3,327 &   4,552 & 3,703 &  6 & 0.2858 \\
PubMed        & 19,717 &  44,324 &   500 &  3 & 0.2658 \\
Amazon PC     & 13,752 & 245,861 &   767 & 10 & 1.0411 \\
Coauthor PHY  & 34,493 & 247,962 & 8,415 &  5 & 0.7784 \\
\bottomrule
\end{tabular}
\end{table}

%% file: tables/table-main-results.tex
\begin{table*}[t]
\centering
\small
\caption{Comparison of external metrics, internal metrics and cluster number across different methods on five datasets.}
\label{tab:main_resluts}
\setlength{\tabcolsep}{3pt} 
\begin{threeparttable}
\begin{tabular}{llcccccc}
\toprule
Dataset & Method & NMI $\uparrow$ & ARI $\uparrow$ & Modularity $\uparrow$ & CHI $\uparrow$ & $Q \times \text{CHI}$  $\uparrow$ & Found-$K$ \\
\midrule
\multirow{6}{*}{\begin{tabular}[c]{@{}l@{}}Cora \\ ($k=7$)\end{tabular}}
  & DAG$^{\dagger}$ & $\underline{0.5171}_{\pm0.0053}$ & {---} & $0.6999_{\pm0.0125}$ & $9.8684_{\pm0.7115}$ & $6.9069$ & $7.4_{\pm1.2}$ \\
  & Neuromap (GIN) & $0.4730_{\pm0.0167}$ & $0.3470_{\pm0.0422}$ & $\underline{0.7905}_{\pm0.0072}$ & $5.6809_{\pm0.5493}$ & $4.4904_{\pm0.4291}$ & $20.0_{\pm2.8}$ \\
  & DeSE & $0.4995_{\pm0.0140}$ & $0.3445_{\pm0.0332}$ & $0.7719_{\pm0.0088}$ & $7.6672_{\pm0.4981}$ & $5.9190_{\pm0.3907}$ & $14_{\pm1.5}$ \\
  & DGCluster$^{\ddagger}$ & $0.4935_{\pm0.0154}$ & $0.3321_{\pm0.0263}$ & $\mathbf{0.8079}_{\pm0.0036}$ & $5.5578_{\pm0.2654}$ & $4.4890_{\pm0.2114}$ & $21.3_{\pm1.6}$ \\
  & GSEC$^{\ddagger}$ & $0.4843_{\pm0.0338}$ & $\underline{0.4004}_{\pm0.0325}$ & $0.7557_{\pm0.0249}$ & $\underline{10.8517}_{\pm1.1496}$ & $\underline{8.1758}_{\pm0.6285}$ & $7.2_{\pm1.3}$ \\
  & CARROT & $\mathbf{0.5698}_{\pm0.0069}$ & $\mathbf{0.5115}_{\pm0.0047}$ & $0.7233_{\pm0.0019}$ & $\mathbf{12.2415}_{\pm0.0446}$ & $\mathbf{8.8559}_{\pm0.0400}$ & $8.0_{\pm0.0}$ \\
\midrule
\multirow{6}{*}{\begin{tabular}[c]{@{}l@{}}CiteSeer \\ ($k=6$)\end{tabular}}
  & DAG$^{\dagger}$ & $\underline{0.4118}_{\pm0.0022}$ & {---} & $0.7435_{\pm0.0194}$ & $\underline{9.1846}_{\pm0.0730}$ & $6.8288$ & $6.4_{\pm0.5}$ \\
  & Neuromap (GIN) & $0.2351_{\pm0.0192}$ & $0.1122_{\pm0.0224}$ & $\underline{0.8498}_{\pm0.0129}$ & $4.0083_{\pm0.2784}$ & $3.4080_{\pm0.2674}$ & $29.4_{\pm3.2}$ \\
  & DeSE & $0.3451_{\pm0.0072}$ & $0.2251_{\pm0.0128}$ & $0.8210_{\pm0.0172}$ & $6.0046_{\pm0.7192}$ & $4.9400_{\pm0.6868}$ & $20.4_{\pm2.9}$ \\
  & DGCluster$^{\ddagger}$ & $0.2571_{\pm0.0088}$ & $0.1308_{\pm0.0134}$ & $\mathbf{0.8726}_{\pm0.0027}$ & $3.4724_{\pm0.1196}$ & $3.0300_{\pm0.1090}$ & $43.8_{\pm3.1}$ \\
  & GSEC$^{\ddagger}$ & $0.3086_{\pm0.0324}$ & $\underline{0.2718}_{\pm0.0360}$ & $0.8128_{\pm0.0144}$ & $8.4759_{\pm1.4222}$ & $\underline{6.8798}_{\pm1.0938}$ & $9.5_{\pm2.1}$ \\
  & CARROT & $\mathbf{0.4320}_{\pm0.0083}$ & $\mathbf{0.4430}_{\pm0.0114}$ & $0.7323_{\pm0.0084}$ & $\mathbf{15.4103}_{\pm0.5736}$ & $\mathbf{11.2833}_{\pm0.3003}$ & $5.3_{\pm0.5}$ \\
\midrule
\multirow{6}{*}{\begin{tabular}[c]{@{}l@{}}PubMed \\ ($k=3$)\end{tabular}}
  & DAG$^{\dagger}$ & $\underline{0.2828}_{\pm0.0143}$ & {---} & $0.5939_{\pm0.0507}$ & $189.2995_{\pm37.5518}$ & $112.4250$ & $3.4_{\pm0.5}$ \\
  & Neuromap (GIN) & $0.2274_{\pm0.0195}$ & $\underline{0.1611}_{\pm0.0285}$ & $\mathbf{0.7317}_{\pm0.0173}$ & $101.6522_{\pm7.9433}$ & $74.3495_{\pm5.5743}$ & $12.5_{\pm2.1}$ \\
  & DeSE & $0.1943_{\pm0.0510}$ & $0.1242_{\pm0.0243}$ & $0.6022_{\pm0.0673}$ & $163.6634_{\pm57.5109}$ & $96.9050_{\pm28.0849}$ & $8.4_{\pm3.4}$ \\
  & DGCluster$^{\ddagger}$ & $0.2466_{\pm0.0148}$ & $0.1445_{\pm0.0185}$ & $\underline{0.7283}_{\pm0.0154}$ & $130.4143_{\pm14.2770}$ & $95.1030_{\pm11.4420}$ & $12.1_{\pm2.2}$ \\
  & GSEC$^{\ddagger}$ & $0.1560_{\pm0.0470}$ & $0.1283_{\pm0.0478}$ & $0.6057_{\pm0.0130}$ & $\underline{202.0660}_{\pm31.9133}$ & $\underline{122.1383}_{\pm18.0933}$ & $3.3_{\pm0.5}$ \\
  & CARROT & $\mathbf{0.3023}_{\pm0.0234}$ & $\mathbf{0.2830}_{\pm0.0370}$ & $0.5783_{\pm0.0286}$ & $\mathbf{296.4590}_{\pm13.5858}$ & $\mathbf{175.4279}_{\pm5.9752}$ & $3.4_{\pm0.5}$ \\
\midrule
\multirow{6}{*}{\begin{tabular}[c]{@{}l@{}}Amazon PC \\ ($k=10$)\end{tabular}}
  & DAG$^{\dagger}$ & {---} & {---} & {---} & {---} & {---} & {---} \\
  & Neuromap (GCN) & $\underline{0.5305}_{\pm0.0209}$ & $\underline{0.4108}_{\pm0.0401}$ & $\underline{0.6055}_{\pm0.0191}$ & $43.1247_{\pm4.3425}$ & $26.1555_{\pm3.1374}$ & $19.2_{\pm1.9}$ \\
  & DeSE & $0.4352_{\pm0.0538}$ & $0.2742_{\pm0.0711}$ & $0.5147_{\pm0.0264}$ & $30.0511_{\pm4.5974}$ & $15.4110_{\pm2.0932}$ & $28.6_{\pm4.7}$ \\
  & DGCluster$^{\ddagger}$ & $0.5018_{\pm0.0194}$ & $0.3489_{\pm0.0185}$ & $\mathbf{0.6106}_{\pm0.0097}$ & $\underline{123.6707}_{\pm26.9816}$ & $\underline{75.2760}_{\pm14.8819}$ & $6.8_{\pm1.40}$ \\
  & GSEC$^{\ddagger}$ & $0.4479_{\pm0.0271}$ & $0.3114_{\pm0.0259}$ & $0.5978_{\pm0.0090}$ & $\mathbf{136.5546}_{\pm9.5862}$ & $\mathbf{81.5798}_{\pm5.1331}$ & $4.9_{\pm0.5}$ \\
  & CARROT & $\mathbf{0.5488}_{\pm0.0179}$ & $\mathbf{0.5115}_{\pm0.0217}$ & $0.5363_{\pm0.0131}$ & $80.1607_{\pm4.8663}$ & $42.9822_{\pm2.6542}$ & $9.5_{\pm1.2}$ \\
\midrule
\multirow{6}{*}{\begin{tabular}[c]{@{}l@{}}Coauthor PHY \\ ($k=5$)\end{tabular}}
  & DAG$^{\dagger}$ & {---} & {---} & {---} & {---} & {---} & {---} \\
  & Neuromap (LIN) & $\underline{0.5818}_{\pm0.0369}$ & $\underline{0.3950}_{\pm0.0633}$ & $0.6030_{\pm0.0201}$ & $181.3808_{\pm20.4785}$ & $109.3681_{\pm12.8137}$ & $15.3_{\pm1.8}$ \\
  & DeSE & {---} & {---} & {---} & {---} & {---} & {---} \\
  & DGCluster$^{\ddagger}$ & $0.5545_{\pm0.0190}$ & $0.3036_{\pm0.0271}$ & $\mathbf{0.6753}_{\pm0.0030}$ & $262.3458_{\pm19.7036}$ & $177.1960_{\pm13.5625}$ & $10.8_{\pm1.0}$ \\
  & GSEC$^{\ddagger}$ & $0.5119_{\pm0.0536}$ & $0.3133_{\pm0.0694}$ & $\underline{0.6390}_{\pm0.0131}$ & $\mathbf{389.2470}_{\pm68.1480}$ & $\mathbf{248.2235}_{\pm40.3755}$ & $5.6_{\pm1.1}$ \\
  & CARROT & $\mathbf{0.7194}_{\pm0.0190}$ & $\mathbf{0.8027}_{\pm0.0202}$ & $0.4966_{\pm0.0044}$ & $\underline{380.2277}_{\pm25.0868}$ & $\underline{188.7479}_{\pm11.6863}$ & $6.3_{\pm0.6}$ \\
\bottomrule
\end{tabular}
\begin{tablenotes}[flushleft]
\item[$\dagger$] Results taken directly from the original paper ($K_{\max} = 2K_{true}$); source code is unavailable.
\item[$\ddagger$] Does not require a $K_{\max}$ parameter.
\item All other methods are evaluated with $K_{\max} = 50$.
\item The best and second-best methods are highlighted in \textbf{bold} and \underline{underlined} for each dataset-metric pair.
\item {---} indicates the result is unavailable.
\end{tablenotes}
\end{threeparttable}
\end{table*}

%% file: tables/table-ablation.tex
\begin{table}[ht]
\centering
\caption{Ablation study results.}
\label{tab:ablation_mean}
\resizebox{\columnwidth}{!}{%
\begin{tabular}{llcccccc}
\toprule
Dataset & Variant & NMI $\uparrow$ & ARI $\uparrow$ & Modularity $\uparrow$ & CHI $\uparrow$ & Found-$K$ & Size CV \\
\midrule
\multirow{3}{*}{\begin{tabular}[c]{@{}l@{}}Cora \\ ($k=7$)\end{tabular}}
  & w/o semi-relaxed & 0.4550 & 0.1178 & 0.5723 & 4.1308 & 48.9 & 0.1951 \\
  & w/o $\mathcal{L}_{comp}$ & 0.4728 & 0.1561 & 0.6206 & 4.1312 & 50.0 & 0.5592 \\
  & Full model       & 0.5699 & 0.5116 & 0.7233 & 12.2412 & 8.0 & 0.6598 \\
\midrule
\multirow{3}{*}{\begin{tabular}[c]{@{}l@{}}CiteSeer \\ ($k=6$)\end{tabular}}
  & w/o semi-relaxed & 0.3283 & 0.0857 & 0.7352 & 4.2696 & 49.9 & 0.1245 \\
  & w/o $\mathcal{L}_{comp}$ & 0.3403 & 0.1222 & 0.7814 & 4.3048 & 50.0 & 0.5349 \\
  & Full model       & 0.4320 & 0.4430 & 0.7323 & 15.4103 & 5.3 & 0.1985 \\
\midrule
\multirow{3}{*}{\begin{tabular}[c]{@{}l@{}}PubMed \\ ($k=3$)\end{tabular}}
  & w/o semi-relaxed & 0.2278 & 0.0399 & 0.5185 & 43.9402 & 49.9 & 0.1010 \\
  & w/o $\mathcal{L}_{comp}$ & 0.2348 & 0.0487 & 0.5572 & 45.0753 & 50.0 & 0.4230 \\
  & Full model       & 0.3023 & 0.2831 & 0.5782 & 296.4415 & 3.4 & 0.4070 \\
\midrule
\multirow{3}{*}{\begin{tabular}[c]{@{}l@{}}Amazon PC \\ ($k=10$)\end{tabular}}
  & w/o semi-relaxed & 0.4671 & 0.1042 & 0.4171 & 38.5516 & 48.4 & 0.2142 \\
  & w/o $\mathcal{L}_{comp}$ & 0.4867 & 0.1370 & 0.4828 & 35.9837 & 50.0 & 0.5639 \\
  & Full model       & 0.5488 & 0.5115 & 0.5363 & 80.1607 & 9.5 & 1.0692 \\
\midrule
\multirow{3}{*}{\begin{tabular}[c]{@{}l@{}}Coauthor PHY \\ ($k=5$)\end{tabular}}
  & w/o semi-relaxed & 0.4506 & 0.0988 & 0.4779 & 98.8818 & 42.2 & 0.4084 \\
  & w/o $\mathcal{L}_{comp}$ & 0.4459 & 0.0925 & 0.4915 & 87.7950 & 50.0 & 0.5018 \\
  & Full model       & 0.7194 & 0.8027 & 0.4966 & 380.2310 & 6.3 & 0.8879 \\
\bottomrule
\end{tabular}}
\end{table}

%% file: Sections/6.Conclusion.tex
\section{Conclusion}
\label{sec:conclusion}

\steve{
We presented CARROT, a competitive adaptive graph clustering framework based on semi-relaxed optimal transport that jointly addresses two \steve{long-standing} challenges in deep graph clustering: unknown cluster number $K$ and cluster size imbalance. Its Adaptive Clustering Mechanism integrates semi-relaxed OT, a competition loss, and hard pruning to progressively eliminate redundant cluster prototypes and recover the underlying cluster count without prior knowledge of $K$.

Extensive experiments on five public benchmarks demonstrate that CARROT consistently achieves state-of-the-art NMI and ARI, accurately recovers the number of clusters without supervision, and produces partitions that more faithfully capture size-imbalanced community structures than existing methods. Moreover, CARROT remains robust to the choice of $K_{\max}$, requiring only a loose upper bound rather than careful hyperparameter tuning.

We hope CARROT motivates future research on adaptive clustering frameworks that combine semi-relaxed optimal transport with prototype competition to automatically discover meaningful cluster structures beyond graph-structured data, where unknown cluster numbers and size imbalance remain broadly unresolved.}

%% file: Sections/appendix_A.tex
\section{Implementation Details}
\label{app:reproducibility}

\subsection{Training Configuration}
\label{app:hyperparameters}

Table~\ref{tab:hyperparameters} reports the dataset-specific hyper-parameters used during pre-training and joint optimization. \steve{For Coauthor PHY, the input node features are reduced to 512 dimensions via PCA before being fed into the GNN due to memory constraints.}
\input{tables/table-hyper}

The remaining hyper-parameters are shared across all datasets. The GNN encoder produces 32-dimensional embeddings. Pre-training uses a learning rate of $10^{-4}$, and the Sinkhorn solver runs for 10 iterations per epoch. We set the entropic regularization strength $\epsilon = 0.1$, the unbalanced penalty $\rho = 0.05$, and the dynamic loss weighting ratio $r = 2.0$. The pruning threshold coefficient is set to $\alpha = 0.05$, pruning begins after 200 epochs of joint optimization, and the pruning interval is $T_{\text{prune}} = 20$ epochs. All experiments use $K_{\max} = 50$. Model checkpoints are evaluated every 10 epochs using the $Q \times \text{CHI}$ criterion, with an early-stopping patience of 500 epochs; the checkpoint with the highest $Q \times \text{CHI}$ score is retained as the final model.

\steve{Beyond these hyperparameters, one implementation-level constant is introduced solely for numerical stability and is not treated as a tunable parameter. Specifically, the cost-matrix masking described
in \Cref{subsubsec:method_pruning_masking} sets $C_{ij} \leftarrow 10^{9}$ for all inactive prototypes $j$, effectively forcing the corresponding Sinkhorn kernel entries $K_{ij}$ to zero and preventing mass transport to pruned prototypes.}

\subsection{Computational Environment}
\label{app:environment}

CARROT, DGCluster, Neuromap, and DeSE are trained on an NVIDIA GeForce RTX 2060 GPU with 6 GB of memory. \steve{GSEC is trained on an NVIDIA RTX A6000 GPU with 49 GB of memory following its official implementation. All methods are evaluated over ten runs (random seeds 0--9), and we report the mean performance.}

\section{Time Complexity}
\label{app:complexity}
The per-epoch complexity of CARROT can be decomposed into three main components, where $D$ denotes the input feature dimension and $d$ denotes the embedding dimension. The two-layer GCN encoder costs $O(|\mathcal{E}|D + ND^2)$ per epoch, taking $D$ as an upper bound on the input feature, hidden, and embedding dimensions. Constructing the cost matrix by computing pairwise squared Euclidean distances between L2-normalized node embeddings and prototype vectors requires $O(NK_{\max}d)$. The Sinkhorn solver iterates over the $N \times K_{\max}$ kernel matrix for $T$ iterations, costing $O(TNK_{\max})$; since $T$ is a small constant fixed across all experiments, this reduces to $O(NK_{\max})$, which is absorbed by the $O(NK_{\max}d)$ term of the cost matrix computation. All loss computations are of lower order and do not affect the overall complexity. The total per-epoch complexity is therefore
\begin{equation}
    O\!\left(|\mathcal{E}|D + ND^2 + NK_{\max}d\right).
\end{equation}

%% file: tables/table-hyper.tex
\begin{table}[htbp]
\centering
\caption{Dataset-specific hyper-parameters.}
\label{tab:hyperparameters}
\resizebox{\columnwidth}{!}{%
\begin{tabular}{lcccc}
\toprule
Dataset & Hidden Dim & Pre-training Epochs & Joint Opt.\ LR (Encoder) & Joint Opt.\ LR ($\mathbf{M}$) \\
\midrule
Cora          & 256 & 300 & $10^{-6}$ & $10^{-2}$ \\
CiteSeer      & 256 & 300 & $10^{-6}$ & $10^{-2}$ \\
PubMed        & 128 & 300 & $10^{-6}$ & $10^{-2}$ \\
Amazon PC     & 256 & 500 & $10^{-7}$ & $5\times10^{-3}$ \\
Coauthor PHY  & 128 & 300 & $10^{-6}$ & $10^{-2}$ \\
\bottomrule
\end{tabular}
}
\end{table}

%% file: Sections/appendix_B.tex
\steve{
\section{Details of Datasets}
\label{app:dataset_details}
We evaluate CARROT on five public benchmarks. Cora, CiteSeer, and PubMed are citation networks~\cite{data_sen2008collective}, where nodes represent papers, edges denote citation links, node features are bag-of-words representations of paper abstracts, and node labels correspond to research topics. Amazon PC~\cite{data_shchur2018pitfalls} is a co-purchase graph, where nodes represent products, edges connect frequently co-purchased items, node features are extracted from product reviews, and node labels indicate product categories. Coauthor PHY~\cite{data_shchur2018pitfalls} is a co-authorship graph, where nodes represent authors, edges denote co-authorship, node features are extracted from publication keywords, and node labels correspond to research fields. Table~\ref{tab:datasets} summarizes the dataset statistics.
}

\section{On the Bias of the CHI Metric}
\label{app:CHI_bias}
Given a partition into $K$ clusters over $N$ nodes, the Calinski--Harabasz Index (CHI) can be expressed as
\begin{equation}
    \text{CHI} = \frac{D_{\text{between}}}{D_{\text{within}}} \times \frac{N-K}{K-1},
\end{equation}
where $D_{\text{between}}$ and $D_{\text{within}}$ denote the between-cluster and within-cluster dispersion, i.e., the spread of cluster centroids around the global centroid and the spread of points around their own cluster centroid, respectively. CHI is known to exhibit a systematic bias toward coarser partitions. Compared with other model-selection heuristics such as the elbow rule, \citet{results_chikumbo2019optimal} found that CHI consistently favors fewer clusters, and \citet{results_arbelaitz2013extensive} reported similar observations in a broader benchmark of clustering validity indices. We verify this bias directly by decomposing CHI into $D_{\text{between}}$, $D_{\text{within}}$, and the $(N-K)/(K-1)$ scaling factor on Amazon PC, comparing runs of our own method that converge to different Found-$K$ across random seeds, as well as across methods with different Found-$K$. In both comparisons, the $D_{\text{between}}/D_{\text{within}}$ ratio decreases only marginally as $K$ decreases, whereas the scaling factor increases substantially, indicating that the resulting rise in CHI is driven primarily by the scaling factor rather than by genuine improvement in cluster separation. Consequently, CHI comparisons are not fully reliable when Found-$K$ differs substantially across methods, as observed on Amazon PC.

\section{Sensitivity to $K_{\max}$: Modularity and CHI}
\label{app:sensitivity_kmax}
\begin{figure}[h]
    \centering
    \includegraphics[width=\columnwidth]{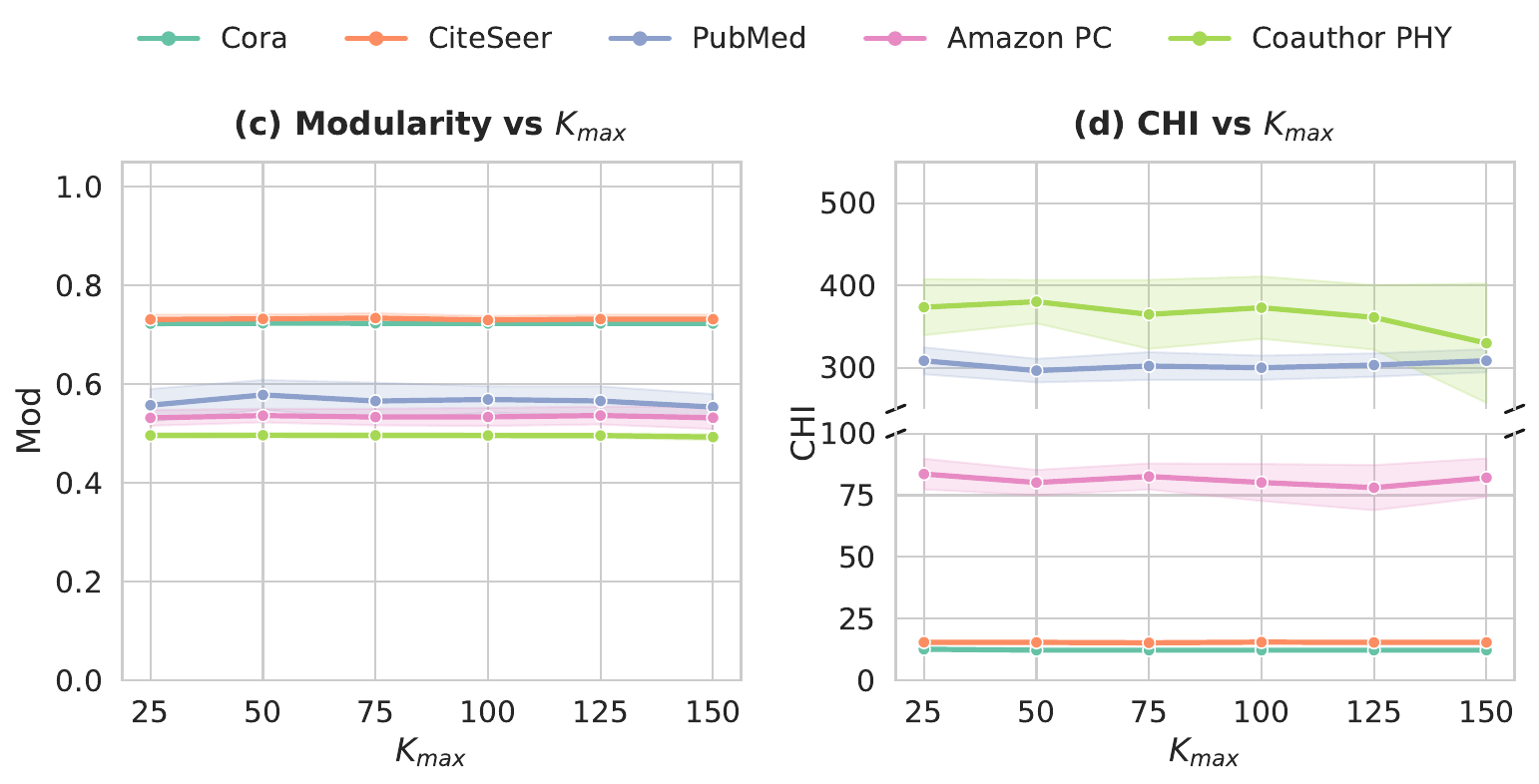}
    \caption{Sensitivity analysis across varying $K_{\max}$: (a) Modularity, (b) CHI. Lines and shaded bands denote mean and standard deviation across seeds.}
    \label{fig:stability_k_appendix}
\end{figure}

\Cref{fig:stability_k_appendix} reports Modularity and CHI for the same sweep of $K_{\max}$ values (25, 50, 75, 100, 125, and 150) described in \Cref{subsec:sensitivity}. Both metrics remain stable across this range on all datasets, consistent with the trends observed for NMI, ARI, and Found-$K$ in the main text. The one exception is CHI on Coauthor PHY at $K_{\max}=150$, which exhibits a modestly larger standard deviation, consistent with the sensitivity of CHI to fluctuations in Found-$K$ noted in Appendix~\ref{app:CHI_bias}.

\section{Sensitivity to $\epsilon$ and $\rho$}
\label{app:sensitivity_eps_rho}
We assess the sensitivity of CARROT to the entropic regularization strength $\epsilon$ and the unbalanced penalty $\rho$, varying one parameter while holding the other fixed at its default value ($\epsilon = 0.1$, $\rho = 0.05$). Table~\ref{tab:eps_rho_sensitivity} reports NMI, ARI, Modularity, CHI, Found-$K$, and Size CV on Cora averaged over 10 random seeds.

\input{tables/table-eps_rho_sensitivity}

CARROT's performance varies considerably with $\epsilon$. NMI and ARI peak at $\epsilon = 0.1$ and decline as $\epsilon$ moves away from this value in either direction, with Found-$K$ falling from $11.0$ at $\epsilon = 0.05$ to $3.7$ at $\epsilon = 1.0$. CHI increases as Found-$K$ decreases, a bias driven predominantly by the CHI scaling factor rather than genuine improvement in cluster separation, as established in Appendix~\ref{app:CHI_bias}. 
This explains why $Q \times \text{CHI}$ peaks at $\epsilon = 0.2$ despite NMI and ARI already being lower than at $\epsilon = 0.1$. At $\epsilon = 0.05$, entropic regularization provides little smoothing and the transport plan concentrates sharply on the nearest prototype, driving aggressive pruning from the earliest epochs and leaving all reported metrics more variable across seeds than at $\epsilon = 0.1$. At $\epsilon = 1.0$, assignments are too diffuse for nodes to develop a clear and consistent preference toward any specific prototype, causing mass to drift continuously across prototypes. This pattern is consistent with $\mathcal{L}_{\text{ot}}$ remaining elevated throughout early and mid training, which, since $\lambda^{(t)}$ scales with $\mathcal{L}_{\text{ot}}$, keeps the dynamic weight disproportionately large as $\mathcal{L}_{\text{comp}}$ gradually rises, sustaining excessive competition pressure until pruning is triggered and cascades rapidly, driving Found-$K$ well below the true partition size. $\epsilon = 0.1$ avoids both failure modes, achieving the highest NMI and ARI with a Found-$K$ standard deviation of zero across all ten seeds.

In contrast, CARROT is markedly robust to $\rho$. Across more than an order of magnitude, from $\rho = 0.005$ to $\rho = 0.2$, NMI varies by only $0.0074$ and ARI by only $0.0062$, well within the run-to-run standard deviation observed at any single setting. Size CV nonetheless decreases monotonically with increasing $\rho$, from $0.6750$ to $0.6003$, consistent with the relaxation factor $\tau = \rho / (\rho + \epsilon)$ approaching $1$ as $\rho$ grows, which tightens the target marginal toward the uniform prior and modestly compresses the recovered size distribution. This trend confirms empirically the role of $\tau$ established in \Cref{subsec:preliminary_srOT}, while the narrow range of NMI and ARI indicates that CARROT's clustering quality does not depend critically on the precise value of $\rho$.

Based on these results, we adopt $\epsilon = 0.1$ for its superior accuracy and stability. For $\rho$, although $\rho = 0.005$ attains marginally higher NMI and ARI, the difference falls within run-to-run noise; we set $\rho = 0.05$ for consistency with the unified hyperparameter setting used across all benchmarks.

%% file: tables/table-eps_rho_sensitivity.tex
\begin{table}[htbp]
\centering
\caption{Sensitivity analysis of $\epsilon$ and $\rho$ on Cora.}
\label{tab:eps_rho_sensitivity}
\resizebox{\columnwidth}{!}{%
\begin{tabular}{llcccccc}
\toprule
Parameter & Value & NMI & ARI & Modularity & CHI & Found-$K$ & Size CV \\
\midrule
\multirow{5}{*}{$\epsilon$}
 & 0.05 & 0.5667 & 0.5026 & 0.7293 & 9.7530  & 11.0 & 0.8819 \\
 & 0.1  & \textbf{0.5698} & \textbf{0.5115} & 0.7233 & 12.2415 & 8.0  & 0.6597 \\
 & 0.2  & 0.5523 & 0.4987 & 0.7168 & 13.5725 & 7.0  & 0.4858 \\
 & 0.5  & 0.4888 & 0.4123 & 0.6439 & 13.3636 & 6.2  & 0.6467 \\
 & 1.0  & 0.3761 & 0.2841 & 0.5375 & 15.6359 & 3.7  & 0.5870 \\
\midrule
\multirow{5}{*}{$\rho$}
 & 0.005 & \textbf{0.5709} & \textbf{0.5128} & 0.7231 & 12.1140 & 8.1 & 0.6750 \\
 & 0.01  & 0.5702 & 0.5122 & 0.7232 & 12.1131 & 8.1 & 0.6739 \\
 & 0.05  & 0.5698 & 0.5115 & 0.7233 & 12.2415 & 8.0 & 0.6597 \\
 & 0.1   & 0.5680 & 0.5085 & 0.7231 & 12.2952 & 8.0 & 0.6478 \\
 & 0.2   & 0.5635 & 0.5066 & 0.7233 & 12.6657 & 7.7 & 0.6003 \\
\bottomrule
\end{tabular}
}
\end{table}